\documentclass[letterpaper, 10 pt, conference]{ieeeconf}

\IEEEoverridecommandlockouts 
\usepackage{amsmath} 
\usepackage{amssymb}  
\usepackage{bm, balance}
\usepackage{algorithm,algpseudocode}
\usepackage[small]{caption}
\usepackage{multicol}
\usepackage{subcaption}
\usepackage{graphicx}
\usepackage{xcolor,mathrsfs,array}
\usepackage[export]{adjustbox}
\usepackage[utf8]{inputenc}
\usepackage[T1]{fontenc}
\usepackage{cite}
\usepackage{flushend}
\usepackage{lipsum}
\usepackage{svg}
\usepackage{booktabs}
\usepackage{multirow}
\usepackage{calc}
\usepackage{makecell}

\usepackage{censor}

\StopCensoring 

\usepackage{cuted}
\usepackage{capt-of}
\usepackage{bm}

\newcommand{\act}{\bm{a}}                 
\newcommand{\nact}{\bm{x}}                
\newcommand{\A}{\mathcal{A}}              
\newcommand{\D}{\mathcal{D}}              
\newcommand{\Pset}{\mathcal{P}}           
\newcommand{\Front}{\mathcal{F}}          
\newcommand{\GP}{\mathcal{GP}}

\newcommand{\obj}{\bm{f}}                 
\newcommand{\nobj}{m}                     
\newcommand{\ndim}{n}                     
\newcommand{\lo}[1]{\underline{a}_{#1}}
\newcommand{\hi}[1]{\overline{a}_{#1}}

\newcommand{\IGDp}{\mathrm{IGD}^{+}}


\usepackage[linkbordercolor={1 1 1},citebordercolor={1 1
  1},urlbordercolor={0.0 0.0
  0.0},urlcolor=blue,colorlinks=true,linkcolor=black,citecolor=black]{hyperref}
\usepackage{url}

\DeclareUnicodeCharacter{2217}{*}

\usepackage{shortcuts} 

\usepackage{pifont}
\usepackage{etoolbox}
\newtoggle{showrevisions}                                             
\togglefalse{showrevisions}

\title{\textbf{Multi-Objective Human-in-the-Loop Bayesian Optimization \\ of a Lower-Limb Exoskeleton}}

\author{Varun Madabushi, Neil Janwani, and Maegan Tucker
    \thanks{This work is supported by the Georgia Tech Institute for Robotics and Intelligent Machines (IRIM) and NSF (CPS Award \#2440387)}
    \thanks{Authors are with the Dynamic Mobility Lab at Georgia Tech, Atlanta, U.S. \texttt{\{vmadabushi, njanwani, mtucker34\}@gatech.edu}}
    }
    
\author{\blackout{Neil Janwani, Matthew T. Lerner, Aaron J. Young, Maegan Tucker}
\thanks{All authors are with the Georgia Institute of Technology.}
\thanks{This work was supported by the National Science Foundation Graduate Research Fellowship Grant No. DGE-2545526 and under IRB H21184.}
}

\begin{document}

\maketitle
\thispagestyle{empty}
\pagestyle{empty}

\begin{abstract}
Human-in-the-loop optimization (HILO) is a common approach for optimizing the control of assistive devices to account for the wearer's unique biomechanics and subjective preferences. However, despite research suggesting that a person may have a different prioritization of objectives depending on time-varying factors such as the environment, their mood, or energy levels, existing HILO approaches only consider a single objective or enforce a fixed weighting on a set of objectives. Neither approach is capable of representing an individual's preferences \textit{over objectives}. In this work, we propose Multi-Objective Human-in-the-loop Bayesian Optimization (MO-HILBO), which builds on explicit multi-objective Bayesian optimization to efficiently infer a personalized set of Pareto-optimal controllers. We compare our approach with an existing multi-objective HILO method and experimentally demonstrate MO-HILBO on a lower-limb exoskeleton across two objectives: metabolic cost (efficiency) and ordinal human feedback (comfort). We find that MO-HILBO (1) discovers Pareto-optimal controllers, and (2) that the pairwise ordering of points on the Pareto front itself is consistent with validation trials. Lastly, we open-source \texttt{mohilo}, a Python package for running both HILO and MO-HILBO on wearable devices: \href{https://dynamicmobility.github.io/mohilo/}{https://dynamicmobility.github.io/mohilo/}.  
\end{abstract}


\IEEEpeerreviewmaketitle

\section{Introduction}

Assistive wearable robots like exoskeletons carry an exciting potential for human augmentation and rehabilitation in everyday tasks including walking \cite{tricomi2024soft, seungmoon2021optimized, zhang2017human, slade2022personalizing}, running \cite{shetty2025ankle}, stair-climbing \cite{park2025human}, and more \cite{kantharaju2022reducing, huo2021impedance, molinaro2024task}. While the benefits of such devices--typically captured by metrics such as metabolic cost, walking speed, gait symmetry, and subjective preference--largely depend on the underlying controller, it is widely recognized that personalization is required to maximize the benefits for each user \cite{slade2022personalizing, tucker2020human, tucker2020preference, ingraham2022role, ingraham2023leveraging,lee2023user,arens2025preference}. Specifically, personalization accounts for factors such as unique gait biomechanics and subjective preferences that vary across individuals. 

\begin{figure}[t!]
    \centering 
    \includegraphics[width=0.9\linewidth]{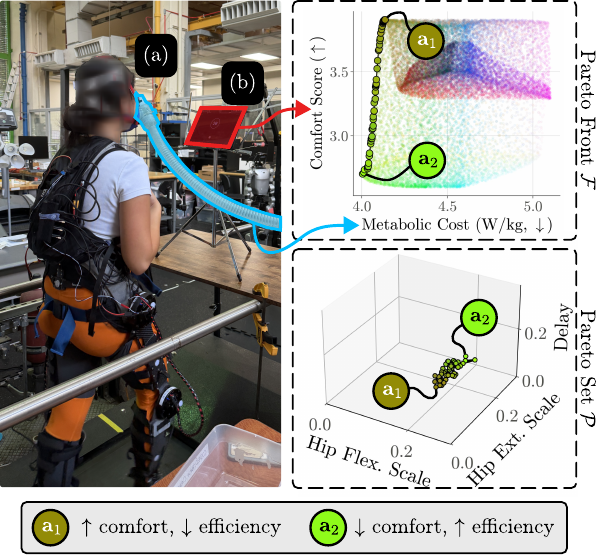} 
    \caption{Our work demonstrates MO-HILBO on a lower-limb exoskeleton (left) over two objectives: (a) metabolic cost, measured by indirect calorimetry, and (b) subjective comfort, recorded using a tablet. The protocol identifies the Pareto set of control parameters (bottom right) defined as the actions associated with the Pareto front, approximated by a learned probabilistic surrogate (top right).}
    \label{fig:hero} 
    \vspace{-6mm}
\end{figure}

Human-in-the-loop Optimization (HILO) is the standard approach for accomplishing personalization \cite{slade2024human}. Broadly, HILO iteratively cycles between 1) selecting new parameters to execute on the device, and 2) collecting corresponding human outcome metrics (e.g. metabolic cost, preference). 
An often overlooked aspect of HILO is selecting an appropriate outcome to optimize. While metabolic cost is the most commonly selected objective \cite{felt2015body, koller2016body, zhang2017human, kim2017human, ding2018human, kim2019bayesian, haufe2020human, witte2020improving, poggensee2021adaptation, bryan2021hip, franks2021comparing, gordon2022human, kantharaju2022reducing, kim2022reducing, slade2022personalizing, kantharaju2023framework, park2025human, powell2025optimized}, it can conflict with other measures such as comfort \cite{young2017influence}.  As such, solely optimizing for one outcome may be detrimental to other important objectives that matter to humans.

In this work, we argue that the wearable device community should move towards optimizing for a \textit{combination of objectives}, rather than only for a single one. One example would be speed and efficiency, explored in \cite{stanfordsmohilo}. 
Here, a scalar-valued multi-objective cost function was constructed by assigning fixed weights to each objective, scaling each objective's impact relative to the other. 
By using two feedback mechanisms, one for each objective, the authors used a traditional HILO setup to complete the optimization. 
However, this requires that the wearer's \textit{trade-off} of objectives (i.e., the scalarizing weighting terms assigned to each objective) be known \textit{prior} to the fitting procedure starting. 
\begin{figure*}[!t]
    \centering
    \includegraphics[width=0.95\textwidth]{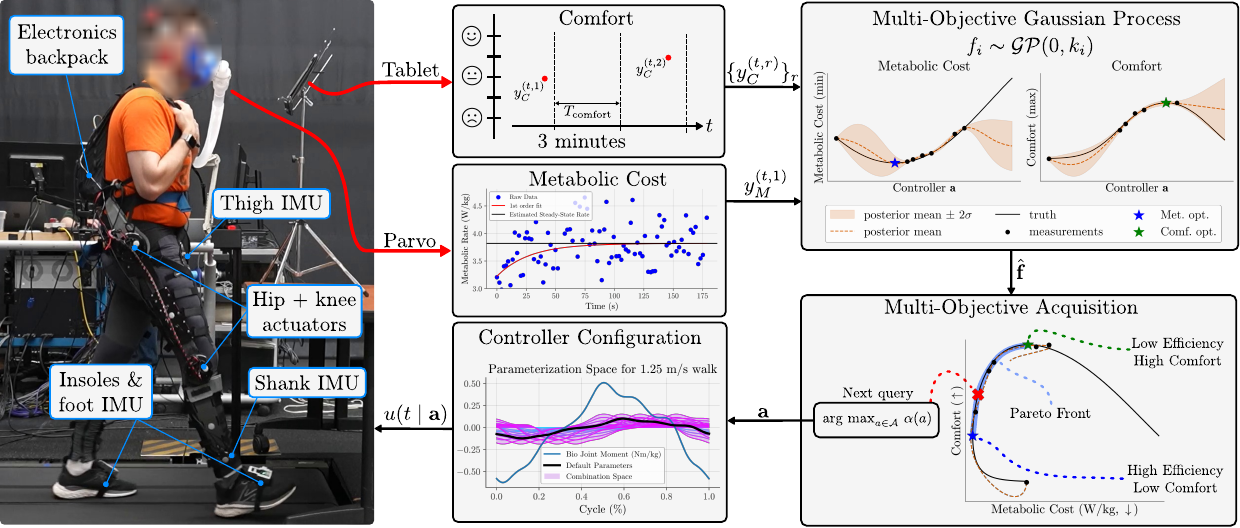}
    \caption{MO-HILBO aims to efficiently identify the Pareto set of mutually-optimal actions across multiple objectives. Similar to traditional HILO, MO-HILBO cycles over three main steps: 1) collecting feedback from human users, 2) updating the belief of the underlying objective function(s) based on the provided feedback, and 3) selecting new queries for the subsequent iteration using an acquisition function. In our work, we explore the application of MO-HILBO on a lower-limb exoskeleton across two objectives: subjective evaluations of comfort using ordinal labels, and metabolic cost measured by indirect calorimetry.}
    \label{fig:methods}
    \vspace{-4mm}
\end{figure*}
Moreover, a wearer's preferred trade-off may not be consistent across individuals (or even over time as the user adapts to the device), as implied by preference-based HILO \cite{tucker2020preference, tucker2020human,ingraham2022role,lee2023user,arens2025preference}, or be consistent in a single individual across different task settings.
For instance, if you are late to a meeting, you might sacrifice efficiency for speed in order to make it on time, despite normally opting for efficiency.  

While canonical optimization minimizes a scalar-valued objective, \textit{multi-objective} optimization considers a vector-valued signal with each entry corresponding to a distinct objective. Since no controller can optimize all objectives simultaneously, the goal is to instead infer a \textit{Pareto-optimal set} of controllers which optimally trade off among objectives. A popular approach that, to our knowledge, has not yet been applied to human assistive robotics, is multi-objective Bayesian optimization (MOBO). MOBO takes a sample-efficient, and noise-aware approach by intelligently querying different controllers to balance exploration and exploitation.

Of particular relevance is the work of \cite{mohilo}, which introduces non-Bayesian multi-objective HILO by using an evolutionary algorithm: Non-dominated Sorting Genetic Algorithm (NSGA-II) \cite{nsga2}. However, with a limited budget, this approach often finds sparse Pareto-optimal points that may not describe the full Pareto set, and must roll out entire generations of parameters to the user before making an update. In comparison, MOBO is a more sample-efficient optimizer that infers dense Pareto sets via a continuous approximation of the objective space. 

In this work, we demonstrate the first, to our knowledge, multi-objective human-in-the-loop Bayesian optimization (MO-HILBO) protocol for wearable devices, as illustrated in Fig.~\ref{fig:hero}. Our explicit contributions are:
\begin{enumerate}
    \item We introduce MO-HILBO and showcase its improved performance on limited sample, high noise regimes compared to previous genetic-based multi-objective HILO approaches.
    \item We develop a MO-HILBO protocol and demonstrate it towards optimizing 3 parameters of a lower-limb exoskeleton controller over two objectives: metabolic cost and subjective comfort.
    \item We introduce and demonstrate a validation procedure for multi-objective HILO, showing that MO-HILBO identifies Pareto-optimal points and that the ordering of the front with respect to a single objective is largely consistent for each user.
    \item We provide \texttt{mohilo}\footnote{\label{fn:code}Project website: \href{https://dynamicmobility.github.io/mohilo/}{https://dynamicmobility.github.io/mohilo/}.}: a Python framework for using both MO-HILBO and single-objective Bayesian Optimization for HILO on assistive robots.
\end{enumerate}





\section{MO-HILBO}
The MO-HILBO framework, illustrated in Fig.~\ref{fig:methods}, has two main components: the multi-objective Gaussian Process (MO-GP) model, which infers the Pareto-optimal controller set, and the multi-objective acquisition function, which picks the next controller to query based on the MO-GP's current estimate of the Pareto set. While this acquisition can be modified for different end goals, we select an acquisition function to efficiently explore the trade-off space, with the goal of identifying a Pareto front with the largest hypervolume as quickly as possible. Our algorithm is open-sourced through the Python package \texttt{mohilo}$^{\ref{fn:code}}$, making it available for practitioners who wish to perform sample-efficient multi-objective HILO.

\subsection{Problem Setup}
The goal of MO-HILBO is to identify a set of actions, with each action $\bm{a} \in \mathcal{A} \subset \mathbb{R}^n$ being a tuple of $n$ tuneable parameters, which optimally trade off across a set of $m$ objectives $\obj : \A \to \mathbb{R}^{\nobj}$ (i.e., $\obj(\act) = \big[f_1(\act), \dots, f_{\nobj}(\act)\big]^{\top}$, with $f_i: \mathcal{A} \to \mathbb{R}$). 
In other words, we wish to optimize these $m$ objectives simultaneously. However, since these objectives may conflict, no single controller is necessarily optimal for all objectives at once. Instead, MO-HILBO solves for the Pareto set of controllers $\Pset_{\obj}$, which
contains only controllers that are \textit{non-dominated}:
%

\begin{equation}
  \Pset_{\obj} := \big\{\act \in \A \;:\; \nexists\, \act' \in \A
  \text{ with } \act' \succ \act \big\},
\end{equation}
where $\succ$ denotes Pareto dominance over the $m$ objectives
$f_1, \dots, f_m$:
%
\begin{equation}
  \act' \succ \act
  \iff
  \begin{cases}
    f_j(\act') \ge f_j(\act) & \forall j \in \{1,\dots,\nobj\}, \\[2pt]
    f_j(\act') >   f_j(\act) & \text{for some } j.
  \end{cases}
\end{equation}

Plainly stated, controller $\bm{a}$ is excluded from $\Pset_{\obj}$ whenever another controller $\bm{a}'$ can perform as well as $\bm{a}$ on every objective and strictly improve on at least one. 
With this, we define the Pareto front $\mathcal{F}$ as
\begin{equation}
  \Front_{\obj} := \obj(\Pset_{\obj}) = \{\obj(\act) : \act \in \Pset_{\obj}\}
  \subset \mathbb{R}^{\nobj}.
\end{equation}

For notational consistency in this section, we seek to \textit{maximize} all objectives. While this is not always the case (e.g., in practice we seek to \textit{minimize} metabolic cost), we use a standardization procedure in MO-HILBO's initialization that transforms all objectives into maximization space.

\subsection{Multi-Objective Gaussian Process}
With finding $\mathcal{P}_{\bm{f}}$ as the express goal of MO-HILBO, it is tempting to simply evaluate $\bm{f}$ on many actions and choose the non-dominated ones as $\mathcal{P}_{\bm{f}}$. However, simply evaluating each objective $f_i$ requires eliciting biomechanical or psychometric feedback from the human, making it a costly operation.
Thus, we instead learn a \textit{probabilistic surrogate} of the true objectives $\bm{f}$ using an independent GP for each objective:
%
\begin{equation}
  f_j \sim \GP\big(0,\, k_j(\cdot,\cdot)\big), \qquad j = 1,\dots,\nobj,
\end{equation}
where each GP in the MO-GP represents a non-parametric distribution over the unknown true objective function $f_j$. We denote $k_j$ as the kernel function which determines the prior covariance of any two actions $\bm{a}$, $\bm{a}'$. Several kernels exist in the Gaussian process literature; we use the automatic relevance determination (ARD) squared-exponential kernel:
\begin{equation}
  k_j(\nact, \nact')
  = \sigma_{f,j}^{2}
    \exp\!\left(-\frac{1}{2}\sum_{i=1}^{\ndim}
    \frac{(x_i - x_i')^{2}}{\ell_{j,i}^{2}}\right),
\end{equation}
for its commonplace use in regression of black-box functions. Here, $\{\sigma_{f,j},~\ell_{j,i}\} \in \mathbb{R}_+$ are hyperparameters, denoting the signal variance and lengthscale, respectively. Intuitively, signal variance defines the expected scale of the objective and lengthscale defines the correlation between two actions. 
We optionally set a minimum lengthscale $\ell_{\textrm{min}}$ that prevents each GP from overfitting by collapsing the lengthscale.

There are two primary benefits of using a Gaussian process as a surrogate model. First, is that we obtain a continuous, easy-to-evaluate, posterior of each objective over the action space. Second, we can obtain the confidence of the surrogate by evaluating the variance of the Gaussian. Specific formulations are provided in the model update section below.

\subsection{Pareto Set Discovery}
As summarized in Algorithm \ref{alg:mo-hilbo}, MO-HILBO begins with an initialization phase, followed by iteratively cycling through three steps: (1) choosing the next set of control parameters to roll out to the user, (2) eliciting the corresponding feedback from the human user, and (3) updating the surrogate model. We repeat this for a fixed budget $T$. In practice, we choose the largest $T$ that does not fatigue the user, maximizing the amount of data collected.

\begin{algorithm}[t]
\caption{MO-HILBO}
\label{alg:mo-hilbo}
\begin{algorithmic}[1]
\Require action space $\mathcal{A}$, objectives $1..m$ with directions $\rho_j$, budget $T$, initialization allotment $T_0$
\State $\mathcal{D}_j \gets \emptyset$ for $j = 1, \dots, m$
\For{$t = 1, \dots, T$}
    \If{$t \leq T_0$}
        \State $\bm{a}^{(t)} \sim \mathrm{Unif}(\mathcal{A})$ \Comment{no posterior yet}
    \Else
        \State $\bm{a}^{(t)} \gets$ \texttt{qLogNParEGO}
    \EndIf
    \State Send $\bm{a}^{(t)}$ to the device
    \State Measure $\{y_j^{(t,r)}\}$ in parallel, including repeats.
    \State Store feedback in $D^{(t)}_j$
    \State Refit $\mu$ and $v$ by maximizing \eqref{eq:mll}
\EndFor
\State \Return ($\mu$ and $v$) and the corresponding Pareto set $\mathcal{P}_{\hat{\bm{f}}}^{(T)}$
\end{algorithmic}
\end{algorithm}

\newsubsec{Initialization}
We initialize the algorithm by setting the bounds on the action space $[\lo{i}, \hi{i}]$ for each dimension $i \in [1,n]$. We instantiate a normalizing transform $\varphi$ which maps $\bm{a} \in \mathcal{A}$ to the unit-box:
\begin{equation}
  \nact = \varphi(\act)
  := \left[\frac{a_i - \lo{i}}{\hi{i} - \lo{i}}\right]_{i=1}^{\ndim}
  \in [0,1]^{\ndim}.
\end{equation}

This is convenient for dealing with actions that span different ranges or exist in different units. Second, given the number of objectives $m$ and whether each is to be maximized or minimized, we instantiate a standardizing transform:
\begin{equation}
  z_j
  = \rho_j \,\frac{y_j - \hat\mu_j}{\hat\sigma_j},
  \quad
  \rho_j =
  \begin{cases}
    +1 & \text{if } f_j \text{ is maximized} \\
    -1 & \text{otherwise}
  \end{cases},
  \label{eq:ytoz}
\end{equation}
where $y_j$ is the feedback measure provided for objective $j$, and $\hat\mu_j$ and $\hat\sigma_j$ are per-objective running mean and standard deviation, respectively, computed from the entire feedback dataset. Superscripts (i.e., $y^{(\cdot)}_j$) will be used later to index per-iteration feedback, but are omitted here for clarity. Lastly, we randomly sample $T_0$ actions to begin the algorithm.

\newsubsec{Feedback Collection}
After initialization, MO-HILBO enters a feedback collection stage where the selected action(s) (in our case, control parameters) is rolled out to the user and their feedback is elicited. Standard MOBO assumes that a single feedback point is obtained per objective per action. We relax this structure into \textit{decoupled feedback collection} by appending to the following dataset:
%
\begin{equation}
  \D_j^{(t)}
  = \Big\{\big(\nact^{(t')},\, y_j^{(t',r)}\big)
    \;:\; t' = 1,\dots,t,\;\; r = 1,\dots,c_j^{(t')}\Big\},
\end{equation}
\begin{equation}
  N_j^{(t)} = \sum_{t'=1}^{t} c_j^{(t')},
  \qquad
  N_j^{(t)} \ne N_{j'}^{(t)} \text{ in general},
\end{equation}
where $t \in \mathbb{N}$ is the current iteration number,
and $r$ indexes $c_j^{(t)}\in \mathbb{N}$ repeated measurements for objective $j$ in trial $t$. For example, for iteration $t$ there could have been three repeated measurements of objective one ($\{y_1^{(t,1)},y_1^{(t,2)},y_1^{(t,3)}\}$, $c_1^{(t)} = 3$) and only one for objective two ($y_2^{(t,1)}$, $c_2^{(t)} = 1$).


We found this parallelized, decoupled scheme to significantly reduce the full data collection procedure's duration as it enables multiple feedback points on one objective to be collected while a different, more time-consuming objective is being evaluated. Additionally, when feedback is noisy, evaluating an objective multiple times on a single action helps improve the quality of the surrogate model.

\newsubsec{Updating the Multi-Objective Gaussian Process}
After collecting feedback, we fit a closed-form GP \cite{williams2006gaussian} to each $\mathcal{D}^{(t)}_j$, with mean and variance, respectively computed as:
\begin{align}
  \mu_j^{(t)}(\nact_\star)
  &= \bm{k}_{j,\star}^{\top}
    \big(K_j + \sigma_{n,j}^{2} I\big)^{-1} \bm{z}^{(t)}_j, \\
  v_j^{(t)}(\nact_\star)
  &= k_j(\nact_\star,\nact_\star)
  - \bm{k}_{j,\star}^{\top}
    \big(K_j + \sigma_{n,j}^{2} I\big)^{-1} \bm{k}_{j,\star},
    \label{eq:mu}
\end{align}
where $\bm{z}^{(t)}_j$ is the standardized feedback and vector $[\bm{k}_{j,\star}]_{(t',r)}$ and Gram Matrix $K_j$ are computed as:
\begin{align}
  [K_j]_{(t',r),(t'',r'')} &= k_j\big(\nact^{(t')}, \nact^{(t'')}\big), \\
  [\bm{k}_{j,\star}]_{(t',r)} &= k_j\big(\nact_\star, \nact^{(t')}\big),
\end{align}
for any algorithm iterations $t', t''$. See that, due to repeated feedback, there may exist repeated values in $K_j$, making it singular. However, because each objective is assumed to be noisy, we assign $\sigma_{n,j}$, a non-zero user-defined prior per-objective variance, that is added to the diagonal of $K_j$ above.
We can now define the surrogate Pareto set $\Pset_{\hat{\obj}}^{(t)} \triangleq \Pset_{\bm{\mu}^{(t)}}$, constructed by the vector-valued mean of the individual GPs:
\begin{equation}
  \bm{\mu}^{(t)}(\nact) = \big[\mu_1^{(t)}(\nact),\dots,\mu_{\nobj}^{(t)}(\nact)\big]^{\top}.
\end{equation}



Lastly, because hyperparameters
$\bm{\theta}_j = \{\sigma_{f,j}^2, \ell_{j,i}, \sigma_{n,j}^2\}$ are not necessarily
known in advance, we estimate them from data, as commonly done in the literature \cite{williams2006gaussian}. However, rather than maximizing the log marginal likelihood $\log p(\bm{z}^{(t)}_j \mid X^{(t)}_j, \bm{\theta}_j)$ alone, we incorporate the practitioner-specified prior on the measurement noise and obtain a maximum a posteriori:
\begin{equation}
    \theta_j^{\star}
  = \argmax_{\theta_j}\;
    \log p(\bm{z}^{(t)}_j \mid X^{(t)}_j, \theta_j) + \log p(\sigma_{n,j}^{2}).
    \label{eq:mll}
\end{equation}
with $X^{(t)}_j$ denoting the normalized training set of actions at iteration $t$ and $\bm{z}^{(t)}_j$ the corresponding feedback standardized by Eqn (\ref{eq:ytoz}).

\newsubsec{Choosing the Next Query}
To select a query/action to execute on the system during iteration $t$, denoted $\bm{a}^{(t)}$, we utilize a multi-objective acquisition function $\alpha: [0,1]^n \rightarrow \mathbb{R}$ that trades off between exploitation and exploration given the collected data. We specifically choose \texttt{qLogNParEGO} as implemented in \texttt{BoTorch} \cite{balandat2020botorch}. This acquisition function is a log-domain, noise-aware form of ParEGO \cite{knowles2006parego, daulton2020qehvi, ament2023logei}, which works well in cases with several objectives and noisy feedback.

First, \texttt{qLogNParEGO} draws a trade-off vector uniformly from the simplex:
\begin{equation}
    \bm{w} \sim \mathrm{Unif}(\Delta^{m-1}), \qquad w_j \geq 0, \;\; \textstyle\sum_j w_j = 1,
    \label{eq:weights}
\end{equation}
and then collapses the vector-valued problem into a scalar one with an
\emph{augmented Chebyshev} scalarization:
\begin{equation}
  s_{\bm{w}}(\bm{u})
  = \min_{j}\big(w_j u_j\big)
  + \alpha_{\text{aug}} \sum_{j=1}^{\nobj} w_j u_j,
  \qquad \alpha_{\text{aug}} = 0.05.
\end{equation}
Importantly, $\bm{w}$ is chosen randomly each time $\alpha$ is invoked. This directly enables MO-HILBO to explore the Pareto set through different prioritizations of objectives, rather than choosing a fixed one as in \cite{stanfordsmohilo}.

Second, given $\bm{w}$, \texttt{qLogNParEGO} scores the scalarization candidate using the expected improvement over the best scalarized value of the measured feedback:
\begin{equation}
  \alpha_{\bm{w}}(\nact)
  = \mathbb{E}\Big[
      \big(s_{\bm{w}}(\psi^{(t)}(\hat{\obj}(\nact)))
      - \max_{\nact'}
        s_{\bm{w}}(\psi^{(t)}(\hat{\obj}(\nact')))\big)^{+}
    \Big],
    \label{eq:nei}
\end{equation}
with $(\cdot)^+ := \max(\cdot, 0)$, $\hat{\obj}$ is a posterior sample drawn from the MO-GP given data $\mathcal{D}^{(t)}$, and $\psi^{(t)}: \mathbb{R}^m \rightarrow [0, 1]^m$ is a normalizing transform that ensures relative comparison of objectives on different scales. Lastly, we maximize the acquisition function using \texttt{BoTorch} \cite{balandat2020botorch} to procure the next controller:
\begin{equation}
  \nact^{(t)} = \argmax_{\nact \in [0,1]^{\ndim}} \;\alpha_{\bm{w}}(\nact),
  \qquad
  \act^{(t)} = \varphi^{-1}\big(\nact^{(t)}\big)
\end{equation}
and send $\bm{a}^{(t)}$ to the device, completing the loop.





\section{Synthetic Experiments}

We compare MO-HILBO with Zhang et al. \cite{mohilo}, which is to our knowledge, the only other HILO approach that attempts to approximate the Pareto set of controllers. Zhang et al. utilize the Non-dominated Sorting Genetic Algorithm (NSGA-II) for their core optimization framework. Below, we demonstrate that MO-HILBO results in improved sample efficiency and yields a more accurate approximation of the Pareto front, as indicated by a lower inverted generational distance plus ($\IGDp$) \cite{igdplus}. We evaluate on a test domain that mimics the conditions of a human-subject study: high noise and a small feedback budget.

\begin{figure*}[t!]
    \centering
    \includegraphics[width=\linewidth]{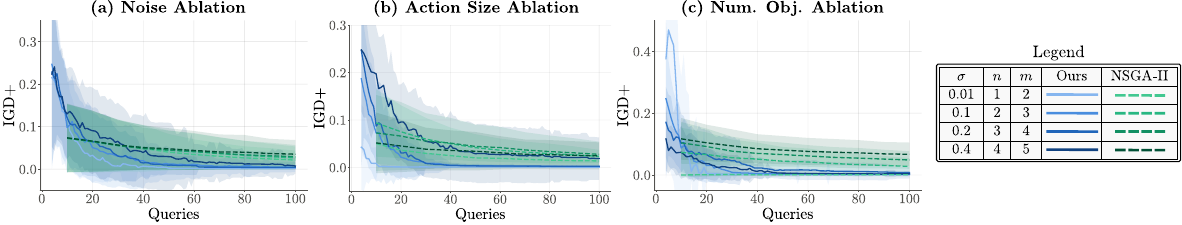}
    \caption{Synthetic experiments on high noise and limited sample budget. Across the ablation studies (left three plots) MO-HILBO achieves a lower $\IGDp$ score across different noise levels, action dimensions, and numbers of objectives than NSGA-II. All curves are averaged over $50$ runs to better measure performance.
    }
    \label{fig:synthetic}
    \vspace{-1mm}
\end{figure*}



\subsection{Comparison Setup}
We construct an ideal-point synthetic function which represents the ground truth objectives. Each objective is optimized at a randomly drawn single action $\bm{a}^*_i \sim \textrm{Unif}(\mathcal{A})$:
\begin{equation}
    f_i(\bm{a}) = - r_i ||\bm{a} - \bm{a}^*_i||_2^2 + \epsilon, \quad \epsilon \sim \mathcal{N}(0, \sigma_{\textrm{syn}}),
\end{equation}
where $r_i \in [0.1, 1]$ is uniformly randomly drawn and $\sigma_{\textrm{syn}}$ is used to inject noise into the synthetic model, representing unavoidable measurement noise in human feedback. When training the GP, we set the prior noise $\sigma_{n,j}$ to within 30\% of the true noise $\sigma_{\textrm{syn}}$ to simulate the realistic case where one does not precisely know the true measurement noise of the groundtruth objective $\bm{f}$.

We also wrap each $f_i(\bm{a})$ in a $\tanh$ function to bound and re-scale the output to $[-1, 1]$. We ablate along three directions: measurement noise ($\sigma_{\textrm{syn}}$), action dimension size ($n$), and number of objectives ($m$). Unless being swept, we set $\sigma_{\textrm{syn}} = 0.1, n = 3, m = 2$, mirroring the action and objective dimensions in our real-world experiment. We perform 50 runs per ablation per algorithm and cap synthetic trials at $T = 100$ to visualize how sample budget affects fit quality and inform our experimental protocol. Lastly, we compute $\IGDp$ using \texttt{pymoo} \cite{pymoo}. 

\subsection{Baseline}
We compare the underlying optimization framework of MO-HILBO (MOBO) against NSGA-II \cite{nsga2, mohilo}. At a high level, NSGA-II sorts a population of sampled points into successive non-dominated fronts and assigns each point a crowding distance that favors points isolated from their neighbors in objective space. Together these define the crowded-comparison operator, which orders points first by front and only then by crowding distance. Offspring are generated from the better-ranked points by crossover and mutation, after which parents and offspring are pooled and truncated back to the population size under the same operator, so non-dominated solutions are never lost (elitism). Over successive generations the population converges to an approximation of the Pareto front. While NSGA-II is very computationally efficient, it does not fit a surrogate model like MOBO does. Thus, approximation only comes through a large number of objective evaluations. For our experiments, we set our population size $P = 10$, matching the size used in the baseline approach \cite{mohilo}.

\subsection{Results}
The results of the synthetic experiments and comparison to NSGA-II are illustrated in Fig.~\ref{fig:synthetic}. Across our ablation study, MOBO reaches better $\IGDp$ scores than NSGA-II with fewer samples, regardless of noise level, action dimension size, or number of objectives. Across $\sigma_{\textrm{syn}}$, MOBO reaches a lower $\IGDp$ score within 40 queries and converges by 100 queries. In contrast, NSGA-II convergence slows as noise increases. Indeed, NSGA-II has no way of distinguishing a falsely high-performing action from a truly non-dominated one once observed. MOBO avoids this problem through its sampling strategy and confidence-aware surrogate model, trialing the same region many times to learn an accurate estimate of the true performance.

On ablating $n$, MOBO runs with $n\in \{1,2, 3\}$ converge quickly. While MOBO $n=4$ outperforms NSGA-II $n=4$, its higher $\IGDp$ score suggests that higher action dimensions are a challenge for MOBO. Lastly, MOBO shows promising performance for all $m\in\{2,3,4,5\}$ objective cases. In contrast, NSGA-II begins to plateau at higher $\IGDp$ values as the number of objectives increases. Overall, these results underscore the regime in which MOBO outperforms evolutionary methods, like NSGA-II: high noise, small action spaces, and limited sample budgets. 


\begin{figure*}
    \centering
    \includegraphics[width=\linewidth]{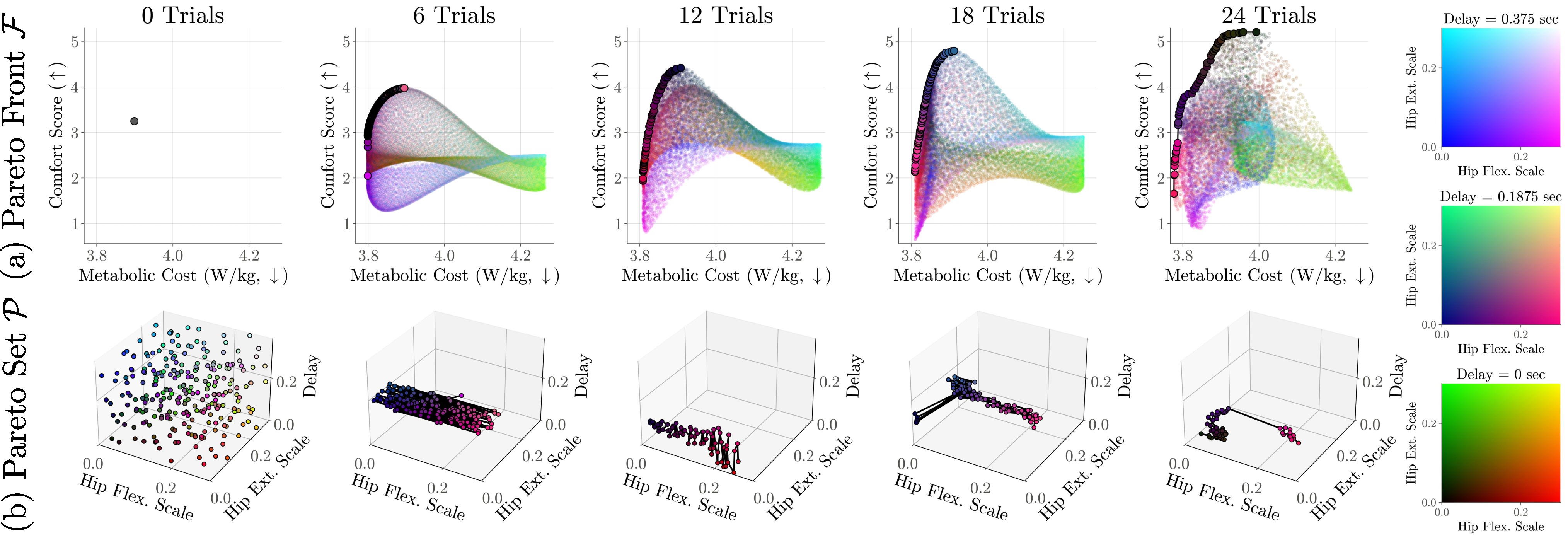}
    \caption{Subject MB03's Pareto set progression throughout the optimization procedure. The Pareto set begins as the entire action space, and eventually clusters around low delay, low hip extension scale. For MB03, hip flexion scale was the primary decision variable for comfort versus efficiency. In fact, MB03's most comfortable gait was when the exoskeleton was turned off, while their most efficient controller was when the hip flexion torque scale was nearly maximized. In short, by performing MO-HILBO, we are able to reduce the search space of controllers from the entire action box, visualized on the bottom left, to a simpler, single-dimension Pareto set, on the bottom right.}
    \label{fig:MB03}
    \vspace{-4mm}
\end{figure*}

\section{Human-Study Experiment}
We construct a real-world human subject study protocol for MO-HILBO and demonstrate optimization of three exoskeleton control parameters across two objectives: metabolic cost (efficiency) and ordinal human feedback (comfort). Moreover, we formulate and carry out a new validation procedure specifically tailored towards multi-objective HILO protocols. 

Using the results from the synthetic study (Fig. \ref{fig:synthetic}), we determined that for $n=3$ and $m=2$, $T = 24$ queries is sufficient to yield informative results while remaining feasible for subjects to complete without fatiguing. To combat high noise observed in the pilot study and a smaller sample budget than ideal, we set the prior noise (in standardized units) to $\sigma_{n,j} = 0.5~\forall j$. We also set $T_0 = 3, \ell_{\text{min}} = 0.1$.

\subsection{Hardware}
We utilize a modified autonomous hip/knee exoskeleton originally developed by \blackout{X, the Moonshot Factory described in Molinaro and Scherpereel et al}. The sensing layer of the device consists of IMUs (3DM-GX5-25, HBK MicroStrain, Williston, VT, USA) embedded in the skin-tight force-transmission leggings, pressure-sensing insoles and foot IMUs placed inside and on the user's shoes (XSENSOR X4, XSENSOR, Calgary, AB, Canada). As described in prior work, the wearable sensor data is mapped to the user's biological joint moment using a temporal convolutional neural network (TCN) for use as the basis of device control \cite{molinaro2024task, molinaro2024estimating}. A single board Linux computer (NVIDIA Jetson Orin Nano, NVIDIA, Santa Clara, CA, USA) manages data acquisition from the sensors, TCN model inference, and communication with the device's four actuators (AK80-9, CubeMars, Jiangxi Province, China). Interfacing with the exoskeleton is handled through an offboard laptop, and participant comfort feedback is collected on an offboard tablet. Both offboard devices are connected to a WiFi network hosted by the onboard microprocessor. The device and mechatronics are pictured in Fig.~\ref{fig:methods}.

\subsection{Exoskeleton Controller Parametrization}
The exoskeleton controller is parametrized by: 
\begin{equation}
    \bm{a}=[\beta_1,\,\beta_2,\,\delta]^{\mathsf T}\in \mathbb{R}^3,
\end{equation}
with $\delta \in [0, 0.375]$ being the temporal delay (in seconds) from the estimated biological joint moment $\tau_{bio}(t-\delta)$ obtained from the TCN, and $\{\beta_{1},\beta_{2}\} \in [0, 0.3]^2$ denoting the scaling coefficients applied during positive and negative biological moment phases, respectively. Given these parameters, the commanded exoskeleton torque is computed as:
\begin{equation} 
u(t \mid \bm{a})= 
    \begin{cases} 
        \beta_{1}\tau_{bio}(t-\delta), & \tau_{bio}(t-\delta) \geq 0, 
\\
        \beta_{2}\tau_{bio}(t-\delta), & \tau_{bio}(t-\delta) < 0. 
    \end{cases}
\end{equation}

 For our study, assistance is applied at the hip while the knee actuators were set to provide zero torque. This reduces the parameter space and therefore the human-subject experiment duration for this initial study of MO-HILBO.


 \begin{figure*}[t!]
     \centering
     \includegraphics[width=\linewidth]{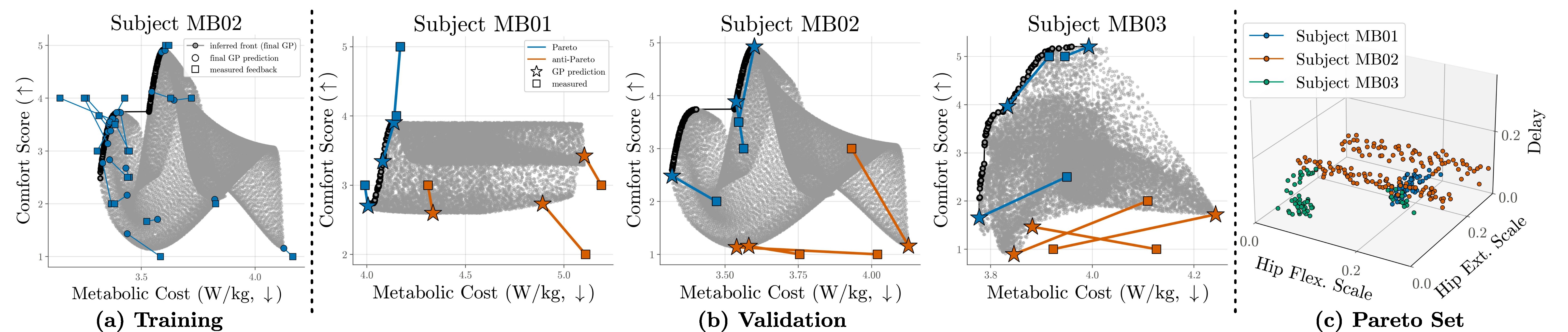}
     \caption{Predicted versus measured points (a), (b) and discovered Pareto sets (c). (a) shows a fitted MO-GP on MB02, where the training data lie near the Pareto front--a result of MO-HILBO's acquisition function exploring and refining the Pareto set. (b) shows the validation points (stars)--which were not seen during training--spaced out over the front. (c) shows the overlayed Pareto sets of all subjects. For the vast majority of points, each subject's Pareto set does not overlap in action space, except for part of MB03's and MB01's. While the subjects' Pareto-optimal controllers featured a variety of flexion or extension torques, none had a high delay.}
     \label{fig:validation}
     \vspace{-4mm}
 \end{figure*}

\subsection{Experimental Design}
Three participants (two males, one female; age: 24.7 $\pm$ 2.3; body mass 75.3 $\pm$ 14 kg; one experienced with the device, two who had not worn the device before) participated in a single-session protocol consisting of MO-HILBO optimization and validation (3-4 hours per subject). Informed consent was obtained from each participant under \blackout{Georgia Institute of Technology Review Board protocol H21184}. 

Metabolic cost and user comfort were collected as participants walked on a treadmill at 1.2 m/s wearing the hip-knee exoskeleton device. Each optimization iteration lasted 3 minutes. Metabolic cost was estimated by fitting a first-order dynamical model to transient metabolic data calculated from oxygen intake and carbon dioxide exhaust measured from each breath (TrueOne 2400, ParvoMedics) using the modified Brockway equation \cite{zhang2017human, brockway1987energy}. User comfort was measured using a tablet (Apple iPad Air) which queried the user of their comfort level on a 1-5 scale (not comfortable to very comfortable) once per minute. We utilized MO-HILBO's decoupled feedback capability to enable the user to answer just one query, or use the remaining two to change their mind, as pilot studies indicated that subjects often changed their minds regarding comfort over the duration of a trial. 

\begin{table}[t!]
\centering
\begin{tabular}{l||c|c||c|c}
\hline
 & \multicolumn{2}{c||}{\textbf{Hypervolume}} & \multicolumn{2}{c}{\textbf{Pairwise Ordering}} \\
\cline{2-5}
Subject
  & \makecell{Pareto}
  & \makecell{Anti-Pareto}
  & \makecell{Metabolic\\Cost}
  & \makecell{Comfort} \\
\hline
MB01 & \textbf{4.02} & 1.300 & 100\% (3/3) & 100\% (3/3) \\
MB02 & \textbf{1.55} & 0.366 & 100\% (3/3) & 100\% (3/3) \\
MB03 & \textbf{1.02} & 0.128 & 67\% (2/3) & 100\% (3/3) \\
\hline
\end{tabular}
\caption{Validation data across subjects. In all cases, the identified Pareto sets were validated to Pareto-dominate the \textit{anti-Pareto} set (built by flipping the sign of each objective). MO-HILBO also correctly predicted 94\% (17/18) of the pairwise orderings of Pareto-optimal points. A valid reference point was chosen for each subject's hypervolume computation, resulting in different scales.
}
\label{tab:validation}
\vspace{-4mm}
\end{table}

\subsection{Validation Protocol}
To evaluate the predictive capability of the learned surrogate model, we validate both the Pareto dominance of the inferred Pareto set and the relative pairwise ranking of its actions. Specifically, we sample three actions from the inferred Pareto set $\mathcal{P}_{\hat{\bm{f}}}$, with two on the endpoints of the front and one in between. We similarly sample three actions from the \textit{anti-Pareto set} $\mathcal{P}_{-\hat{\bm{f}}}$ (found by negating the objectives). These six controllers are evaluated for 5 minutes each by the user in a randomized order during treadmill walking at 1.2~m/s while collecting metabolic cost and user comfort.

Using these validation measurements, we compute two tests: (1) a 3-point estimate of the true hypervolume for both sets to confirm Pareto dominance, and (2) the surrogate's accuracy in predicting the pairwise ordering among the three sampled Pareto actions. For each objective, the surrogate model predicts relative rankings across the three pairwise combinations of these actions. We then compare these predictions against the true measured rankings, yielding an accuracy score based on three binary outcomes per objective.





\section{Results and Discussion}
Our results are shown in Fig.~\ref{fig:MB03} and Fig.~\ref{fig:validation}. For all experiments, we compute $\mathcal{P}_{\hat{\bm{f}}}$ by evaluating the MO-GP mean on a dense sample of the action space. Fig.~\ref{fig:MB03} showcases a single optimization procedure for subject MB03. MO-HILBO began by considering all controllers $\bm{a} \in \mathcal{A}$ as Pareto-optimal (bottom left), and iteratively inferred more selective Pareto sets over decoupled feedback updates (bottom right). In addition to identifying the surrogate Pareto set $\mathcal{P}_{\hat{\bm{f}}}$, MO-HILBO also learns a posterior over all $\bm{a} \in \mathcal{A}$. This is insightful for understanding what the subject \textit{didn't} like. For instance, according to Fig. \ref{fig:MB03}, MB03 found controllers with high torque and delay uncomfortable, and ones with high delay and flexion torque as inefficient.

A secondary goal of this study was to discover if the Pareto sets shared structure across subjects. Qualitatively, the answer is mixed. All subjects found high delay to be neither comfortable nor efficient. However, MB02 and MB03 found low torque as most comfortable, while MB01 found high flexion and extension torque as both comfortable and efficient. Of interest is that MB02 and MB03 had no prior experience in the exoskeleton device, while MB01 did have prior experience. This suggests that there could be non-stationary factors which influence the location of $\mathcal{P}_{\hat{\bm{f}}}$ in addition to other subject characteristics, like physical embodiment and gait style. This is consistent with findings from prior work that demonstrate substantial improvements in metabolic outcomes with training \cite{poggensee2021adaptation}. Thus, an interesting area of future work would be to investigate how these Pareto fronts evolve over extended user exposure and adaptation.

Fig.~\ref{fig:validation} and Table~\ref{tab:validation} summarize the experimental validation results. Across all subjects, the controllers in the surrogate Pareto set $\mathcal{P}_{\hat{\bm{f}}}$ identified by MO-HILBO empirically dominated those from the anti-Pareto set $\mathcal{P}_{-\hat{\bm{f}}}$. The surrogate also accurately predicted the relative pairwise ordering across nearly all validation Pareto actions for both objectives. These were novel controllers not seen during training, indicating that MO-HILBO successfully learns a continuous, probabilistic surrogate of the true Pareto set $\mathcal{P}_{\bm{f}}$. Effectively, MO-HILBO mapped the 3-dimensional action space into a 1-dimensional manifold of controllers that optimally trade off among the objectives. This reduced representation offers a significantly simplified search space for real-time human-in-the-loop tuning, an exciting future research direction.

\section{Conclusion}
This paper introduces the first application of Multi-Objective Human-in-the-Loop Bayesian Optimization (MO-HILBO) for wearable assistive devices. By leveraging probabilistic surrogate models, MO-HILBO improves sample-efficiency compared to prior multi-objective HILO techniques. Through both synthetic benchmarking and lower-limb exoskeleton trials across metabolic cost and user comfort, we demonstrate that MO-HILBO efficiently maps a continuous Pareto set within a practical trial budget ($T = 24$). Further, we construct a validation procedure for MO-HILBO and use it to confirm that MO-HILBO reliably identifies Pareto-dominant controllers and accurately predicts relative pairwise rankings along the learned Pareto front. To support broader adoption, we open-source \texttt{mohilo}$^{\ref{fn:code}}$, a Python package for deploying sample-efficient HILO and MO-HILBO algorithms on wearable systems.

There are several exciting future directions. First, standard Bayesian optimization assumes a stationary objective, which may not capture human adaptation, fatigue, or shifting preferences; future work could model these dynamics as distinct regions of the Pareto set. Second, in cases with many objectives, interactive search algorithms may be required to assist users in navigating large Pareto sets, ensuring tractable, user-centered control.

\bibliographystyle{IEEEtran}
\bibliography{references}

\end{document}